\documentclass[10pt, conference]{ieeeconf}
\IEEEoverridecommandlockouts 

\usepackage{amsmath,amssymb,amsfonts}
\usepackage{graphicx, stfloats}
\usepackage{pifont}

\usepackage{xcolor}
\usepackage[figuresright]{rotating}

\graphicspath{{/}{fig/}}

\usepackage{array}
\usepackage{textcomp}
\usepackage{multirow}
\usepackage{booktabs}

\usepackage{mathtools}
\usepackage{breqn}
\usepackage{float}

\usepackage{pgfplots}
\pgfplotsset{compat=1.7}
\usepgfplotslibrary{groupplots}

\usepackage{subcaption}

\usepackage{multirow,tabularx}
\usepackage{hyperref}
\usepackage{flushend}
\usepackage{algorithmic}
\usepackage[vlined, ruled, shortend]{algorithm2e}

\usepackage{multicol}
\usepackage{times}
\usepackage{cleveref}
\usepackage{soul}
\DeclareMathOperator*{\argmin}{arg\,min}

\usepackage{cuted}

\newlength\figureheight
\newlength\figurewidth
\SetAlCapNameFnt{\footnotesize}
\SetAlCapFnt{\footnotesize}

\newcommand{\eg}{\textit{e.g.,}}%

\crefname{figure}{Fig.}{Figs.}
\Crefname{figure}{Fig.}{Figs.}
\Crefname{section}{Section}{Section}
\crefname{section}{Section}{Section}
\Crefname{table}{Table}{Table}
\crefname{table}{Table}{Table}

\usepackage{etoolbox}
\usepackage{balance}

\title{
    \Huge
    Rheos: Modelling Continuous Motion Dynamics in Hierarchical 3D Scene Graphs
}

\author{
    Iacopo Catalano, Francesco Verdoja, Javier Civera, Jorge Peña-Queralta\authorrefmark{2}, Julio A. Placed\authorrefmark{2}
    \thanks{\authorrefmark{2} Equal Project Management.
    This work was partially supported by the Finnish Cultural Foundation, by DGA\_FSE T73\_23R, and by the Research Council of Finland (decision 354909).
    I. Catalano is with the University of Turku, Finland.
    J. Pe\~na-Queralta is with the Centre for Artificial Intelligence, Zürich University of Applied Sciences, Winterthur, Switzerland.
    J. A.~Placed is with the Instituto Tecnol\'ogico de Arag\'on (ITA) and the University of Zaragoza, Spain.
    J. Civera is with the University of Zaragoza, Spain.
    F. Verdoja is with the School of Electrical Engineering, Aalto University, Finland.
    Corresponding: \texttt{\small imcata@utu.fi}
    }
}

\begin{document}

\maketitle

\begin{abstract}
3D Scene Graphs (3DSGs) provide hierarchical, multi-resolution abstractions that encode the geometric and semantic structure of an environment, yet their treatment of dynamics remains limited to tracking individual agents. Maps of Dynamics (MoDs) complement this by modeling aggregate motion patterns, but rely on uniform grid discretizations that lack semantic grounding and scale poorly. We present Rheos, a framework that explicitly embeds continuous directional motion models into an additional \textit{dynamics} layer of a hierarchical 3DSG that enhances the navigational properties of the graph. Each dynamics node maintains a semi-wrapped Gaussian mixture model that captures multimodal directional flow as a principled probability distribution with explicit uncertainty, replacing the discrete histograms used in prior work. To enable online operation, Rheos employs reservoir sampling for bounded-memory observation buffers, parallel per-cell model updates and a principled Bayesian Information Criterion (BIC) sweep that selects the optimal number of mixture components, reducing per-update initialization cost from quadratic to linear in the number of samples. Evaluated across four spatial resolutions in a simulated pedestrian environment, Rheos consistently outperforms the discrete baseline under continuous as well as unfavorable discrete metrics. \footnotesize{\url{https://github.com/IacopomC/rheos}}

\end{abstract}
\IEEEpeerreviewmaketitle


\section{Introduction}\label{sec:introduction}

\begin{figure}[t]
    \centering
    \includegraphics[width=\linewidth]{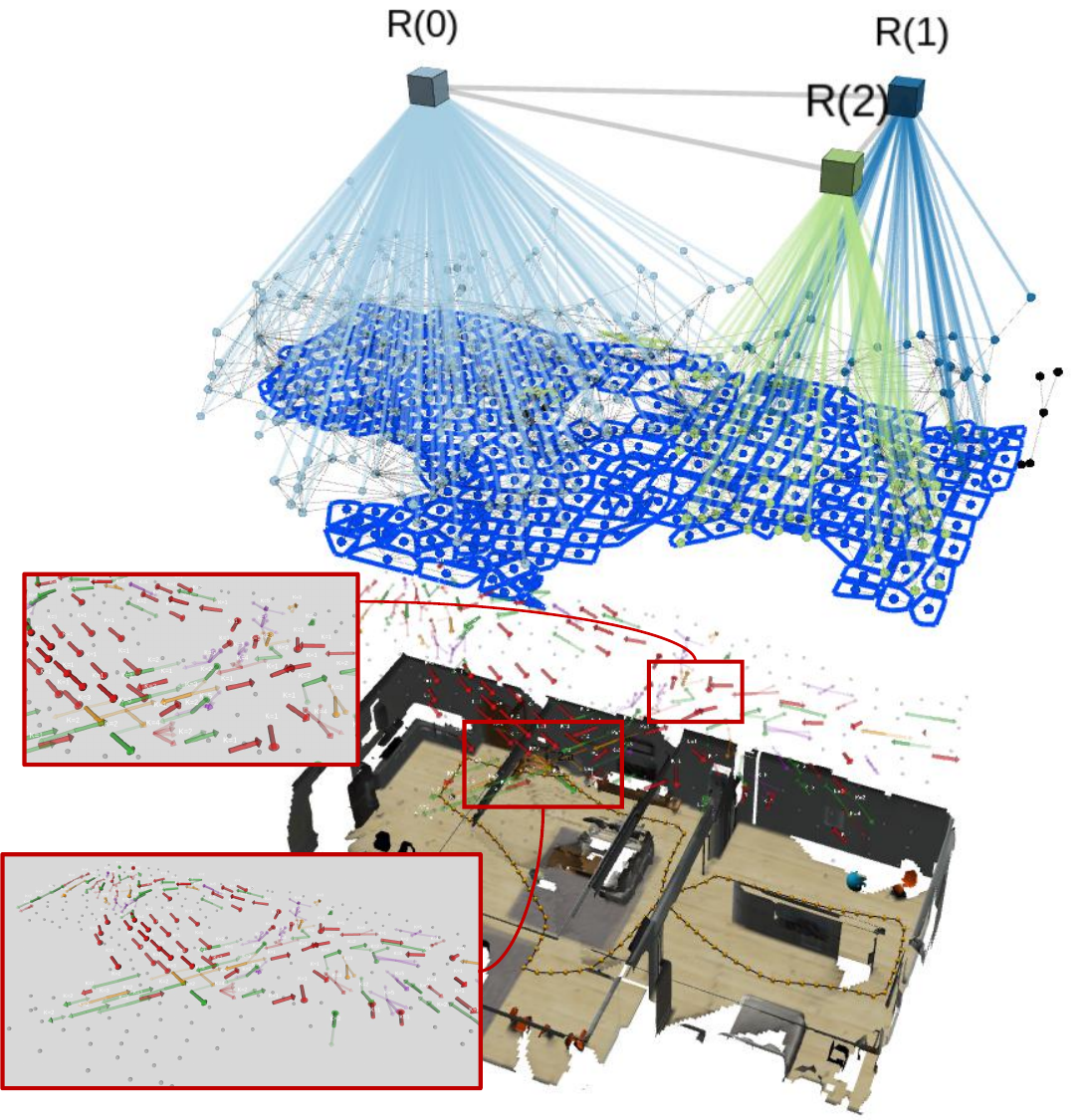}
    \caption{Rheos integrates an environment dynamics layer into a 3DSG through semi-wrapped Gaussian mixture models. Arrow direction represents the component mean angle $\mu^{}_{\theta}$, length is proportional to the mean radial speed $\mu^{}_{\rho}$, and thickness and opacity are scaled by the mixture weight $\alpha^{}_k$. The number of mixture components $K$ is encoded by color ($K = 1$ red, $K = 2$ green, $K = 3$ yellow and $K=4$ purple).\vspace{-1em}}
    \label{fig:rheos}
\end{figure}

Robots operating in human-populated environments must reason beyond static obstacle avoidance to account for the underlying spatiotemporal patterns of human activity. While geometric maps provide the necessary constraints for collision-free planning, they fail to capture the latent \textit{flow} of a scene, such as identifying high-traffic areas during peak hours or recognizing unused shortcuts. Incorporating this knowledge is essential for generating trajectories that are socially compliant and operationally efficient~\cite{grzeskowiak2021crowd, paez2022pedestrian}.

3D Scene Graphs (3DSGs) have emerged as a powerful framework for spatial reasoning, representing environments through hierarchical abstractions that link geometry, semantics, and topology~\cite{catalano20253d}. By organizing space into a hierarchy (\eg~rooms encapsulating navigable regions, which in turn are composed of geometric primitives), 3DSGs provide a semantically grounded discretization of the environment. However, most 3DSG implementations remain largely static. While previous works have integrated dynamic object tracking~\cite{rosinol2020dsg, greve2023curb}, they focus on tracking the transient states of individual agents rather than modeling the aggregate, long-term motion patterns that are characteristic of a specific location.

In parallel, Maps of Dynamics (MoDs) address this by providing probabilistic models of motion learned from repeated observations~\cite{kucner2023survey}. Existing MoD research predominantly utilizes uniform grid maps, employing spectral or probabilistic methods to model temporal periodicity~\cite{krajnik2014spectral, kucner2017enabling}. While computationally convenient, grid-based representations treat all cells uniformly, ignoring the semantic relevance of different areas and remaining disconnected from the hierarchical structures useful for high-level task planning.

This paper introduces \textbf{Rheos} (Greek for ``\textit{flow}"), a framework that integrates continuous directional motion modeling into the navigational layer of a 3DSG. Instead of distributing motion statistics across a rigid grid, Rheos assigns motion models to meaningful locations defined by the scene graph’s spatial hierarchy. As an agent/robot perceives the environment, each node populates a probabilistic model of directional flow. This approach yields a representation that is inherently sparse, interpretable, and topologically grounded, enabling robots to predict location-specific motion dynamics and incorporate them into higher-level planning and navigation.

Our work builds upon Aion~\cite{catalano2025aion}, which first demonstrated the utility of combining MoDs with 3DSGs using discrete orientation histograms. While Rheos retains Aion’s scalable graph architecture and hashing mechanisms, it introduces a fundamental shift in how directional data is represented. We replace discrete histograms with a continuous Semi-Wrapped Gaussian Mixture Model (SW-GMM)~\cite{kucner2017enabling} at each node. This transition addresses the primary limitations of discrete representations: histograms are constrained by fixed angular resolutions and lack a formal mechanism to represent directional uncertainty. By employing SW-GMMs, Rheos captures complex, multimodal directional flows as principled probability distributions, providing explicit uncertainty estimates and a higher-fidelity description of 
environment dynamics.

Furthermore, the formulation in~\cite{kucner2017enabling} from which the SW-GMM model originates is designed for offline batch processing of completed trajectory datasets, requiring a full $\mathcal{O}(n^2)$ mean-shift clustering pass for every model update, and relying on kernel bandwidth to implicitly determine the number of mixture components without a formal model selection criterion.
Rheos adapts this pipeline for online incremental operation by replacing mean-shift initialization with a Bayesian Information Criterion (BIC)-based sweep over candidate component counts that provides a statistically principled choice of model order while reducing per-update initialization from quadratic to linear $\mathcal{O}(n)$ complexity via $K$-means++ seeding, and by introducing reservoir sampling that provides a fixed-memory sample buffer maintaining statistical representativeness under unbounded observation horizons.

The main contributions of this work are:
\begin{itemize}
    \item \textbf{Continuous directional motion modeling in 3DSGs:} We replace the discrete histograms used in prior work with semi-wrapped Gaussian mixture models, providing resolution-independent directional distributions with explicit uncertainty at each navigational node. With Rheos, we introduce for the first time a continuous flow dynamics representation in 3D Scene Graphs.
    \item \textbf{Online incremental model fitting:} We extend the batch fitting procedure in~\cite{kucner2017enabling} for online operation through reservoir sampling for bounded-memory observation buffers and a Bayesian Information Criterion sweep over candidate component counts, yielding a statistically grounded choice of the number of mixture components that is independent of any kernel bandwidth parameter.
\end{itemize}

Overall, we provide the first 3DSG approach that incorporates continuous flow dynamics as an explicit graph layer representation. We also open-source our code for real-time deployment and simulation, and quantitatively evaluate the performance improvements with respect to previous discrete representations across multiple spatial resolutions.
\section{Related Work}\label{sec:related}

\subsection{3D Scene Graphs (3DSGs)}

3DSGs provide a hierarchical abstraction of space, linking geometric primitives and objects to high-level semantic regions through edges that represent spatial proximity and topological containment~\cite{catalano20253d, chang2023hydra, hughes2024foundations}. While initially developed for indoor environments, the formalism has been extended to large-scale urban~\cite{greve2023curb, deng2024opengraph} and unstructured (~\eg agricultural) settings~\cite{mukuddem2024osiris}.

The majority of existing 3DSG frameworks assume a quasi-static topology. While nodes and edges may be incrementally added during exploration, the representation typically lacks a model of motion in the environment. Extensions focusing on affordances and task-oriented reasoning~\cite{ravichandran2022hierarchical, agia2022taskography} annotate nodes with action-relevant attributes, but these properties remain time-invariant.

Research into dynamic environments has primarily focused on tracking discrete entities. Kimera-DSG~\cite{rosinol2020dsg} and subsequent iterations~\cite{rosinol2021kimera} integrate moving agents as dynamic nodes within the static hierarchy, while CURB-SG~\cite{greve2023curb} incorporates dynamic vehicles for urban navigation. While useful, the complexity of these methods scales with the number of observed agents. Aion~\cite{catalano2025aion} represents the closest precursor to Rheos, embedding per-node directional motion histograms within a 3DSG. However, Aion’s reliance on discrete histograms introduces quantization errors and lacks a principled method for representing directional uncertainty. In contrast, Rheos maintains statistics at the node level, characterizing the aggregate directional activity of a location in a continuous manner rather than discretizing it in predefined directions.

\subsection{Maps of Dynamics (MoDs)}

MoDs shift the focus from agent tracking to modeling motion as a continuous property of the environment~\cite{kucner2023survey}. Early formulations utilized pairwise transition probabilities between adjacent grid cells~\cite{kucner2013conditional}, while more recent methods employ per-cell distributions to represent multimodal directional or velocity flows~\cite{kucner2017enabling, senanayake2017bayesian}.

Motion modeling in MoDs often treats each location as a periodic process, using spectral decomposition to identify cycles in occupancy and flow over varying time horizons~\cite{krajnik2017fremen, vintr2019time}. While recent work has explored geometric priors to regularize these learned motion fields~\cite{verdoja2024bayesian} or directly learning how motion patterns evolve from egocentric views~\cite{catalano2026egomod}, a significant limitation remains: existing MoDs almost exclusively utilize uniform spatial grids. This approach assigns the same representational capacity to all areas regardless of their semantic importance or structural complexity. By indexing motion models to the navigational nodes of a 3DSG, Rheos ensures that the spatial granularity of the motion model is aligned with the semantic and topological structure of the scene.
\section{Methodology}\label{sec:methodology}

Rheos addresses the core challenge of integrating continuous flow dynamics within hierarchical 3D scene graphs, yielding a representation that augments 3DSGs with behavioral information about how people move, and enabling consistent dynamics state representation over dynamic graph structures (\cref{fig:rheos}).

\subsection{3D Scene Graphs}

Formally, a hierarchical 3DSG at time $t$ is a graph {$\mathcal{G}^t \triangleq (\mathcal{V}^t, \mathcal{E}^t)$}, where the node set {$\mathcal{V}^t$} is partitioned into $L$ disjoint layers {$\mathcal{V}^t = \bigsqcup_{\ell=1}^{L} \mathcal{V}^t_\ell$}, each corresponding to level of abstraction, with {$\boldsymbol{v}^t_{\ell,i} \in \mathcal{V}^t_\ell$} denoting the i-th node at level $\ell$~\cite{catalano20253d}. Edges {$\mathcal{E}^t$} encode spatial and semantic relationships both within and across layers. In practice, layers typically progress from low-level geometry (\eg~meshes, point clouds) through navigational abstractions (\eg~traversable regions, connectivity) to semantic groupings (\eg~objects, rooms) and global structure (\eg~buildings, cities).

Rheos extends this hierarchy as in~\cite{catalano2025aion} by introducing a dedicated \emph{dynamics layer} at the navigational level. Each navigational node is augmented with a directional flow model that accumulates motion observations over time, characterizing the typical activity at that location. This layer provides the representational foundation for the motion modeling and spatial indexing mechanisms described in the following sections.

\subsection{Sparse Spatial Hashing and Dynamics Binding}

\begin{figure*}[t]
    \centering

    \includegraphics[width=\linewidth]{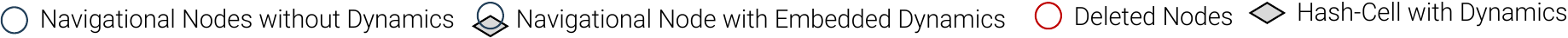}

    \begin{subfigure}[b]{0.24\linewidth}
        \centering
        \includegraphics[width=0.77\linewidth]{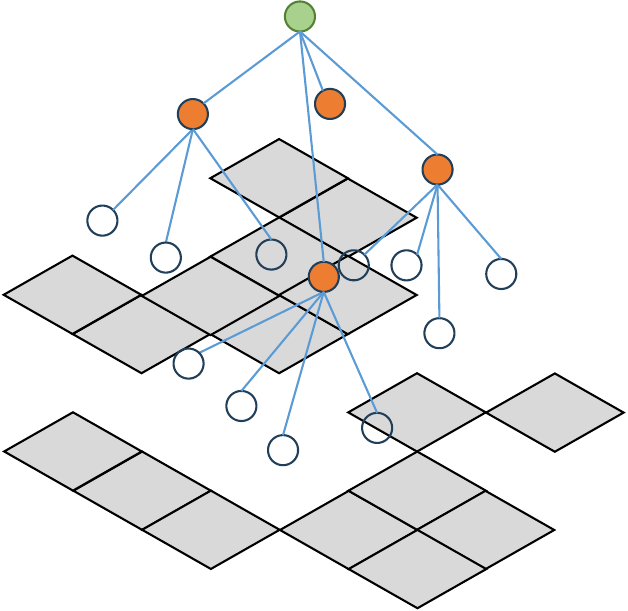}
        \caption{Hash-space accumulation}
        \label{fig:dynamics_in_cell}
    \end{subfigure}\hfill
    \begin{subfigure}[b]{0.24\linewidth}
        \centering
        \includegraphics[width=0.7\linewidth]{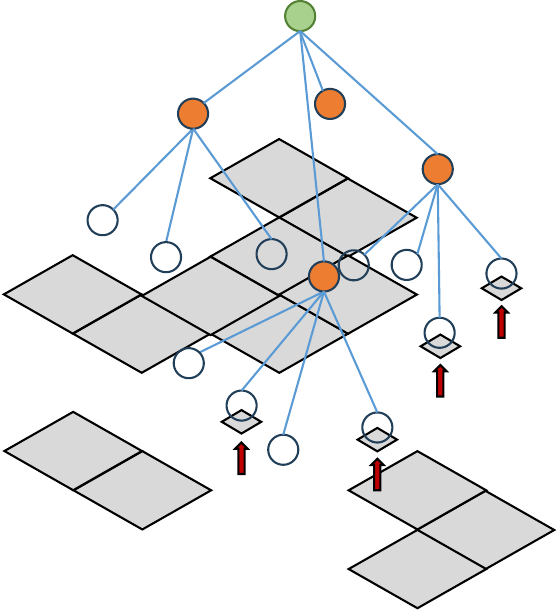}
        \caption{Node binding}
        \label{fig:dynamics_node}
    \end{subfigure}\hfill
    \begin{subfigure}[b]{0.24\linewidth}
        \centering
        \includegraphics[width=0.7\linewidth]{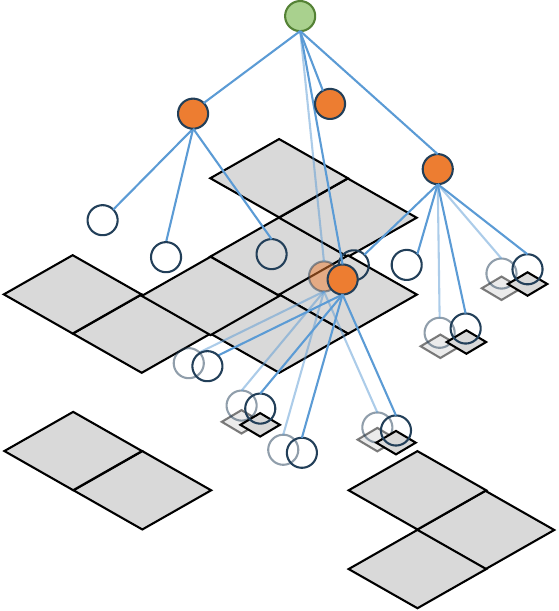}
        \caption{Pose-graph correction}
        \label{fig:dynamics_move}
    \end{subfigure}\hfill
    \begin{subfigure}[b]{0.24\linewidth}
        \centering
        \includegraphics[width=0.7\linewidth]{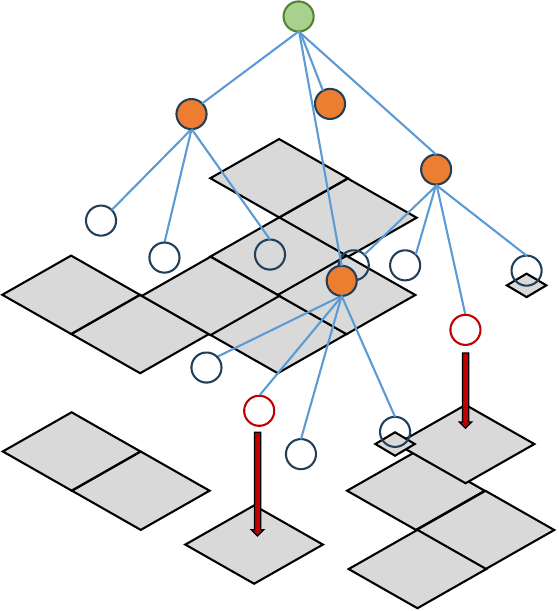}
        \caption{Node removal and reversion}
        \label{fig:dynamics_back}
    \end{subfigure}

    \caption{\textbf{Dynamics ownership lifecycle} (adapted from~\cite{catalano2025aion}). (a)~Motion observations accumulate in hash cells indexed by spatial position. (b)~After the pose graph stabilizes, cell dynamics are transferred to the nearest navigational node. (c)~On loop closure, dynamics move with their owning node. (d)~If a node is removed, its dynamics revert to hash space for future reassignment.}
    \label{fig:spatial_hashing}

\end{figure*}
Rheos inherits from Aion~\cite{catalano2025aion} two infrastructure mechanisms that decouple motion model storage from both fixed grid boundaries and the evolving graph topology.

\textbf{Spatial Hashing.} 
Rather than pre-allocating memory over a bounded grid, motion models are indexed via a sparse hash map (see~\cref{fig:spatial_hashing}). A 3D position {$\boldsymbol{p} = [x, y, z]\in\mathbb{R}^3$} is mapped to a discrete cell key by

\begin{equation}
    h(\boldsymbol{p}) = \text{hash}\left(\left\lfloor \frac{x}{\delta} \right\rfloor, \left\lfloor \frac{y}{\delta} \right\rfloor, \left\lfloor \frac{z}{\delta} \right\rfloor\right)\,,
\end{equation}

\noindent where $\delta$ is the spatial resolution. Memory is allocated only when a cell receives an observation, so storage scales with the number of visited locations rather than with map extent, and lookup remains $\mathcal{O}(1)$.

\textbf{Dynamics Binding.} 
As the pose graph evolves, loop closures may reposition or remove navigational nodes. To prevent motion histories from becoming inconsistent with the current graph state, a binding function maps each hash cell to its owning navigational node. During active exploration, observations accumulate in hash space. Once the graph has stabilized, defined by a window $\tau$ with no significant pose updates, the dynamics state is transferred to the nearest node (\cref{fig:dynamics_in_cell}).
If a node is subsequently repositioned, its dynamics move with it (\cref{fig:dynamics_move}). If it is removed, ownership reverts to hash space (\cref{fig:dynamics_back}) so that the history can be reassigned when a replacement node appears nearby. This transfer scheme prevents duplication and maintains consistency under graph corrections without additional memory overhead.

\subsection{Motion Representation}
\label{sec:flow-modelling}

To model directional dynamics over time, and inspired by~\cite{kucner2017enabling}, Rheos adopts a continuous SW-GMM representation, extended to operate incrementally over navigational nodes in a sparse spatial hash rather than over a fixed grid.

For notational simplicity, let us denote by {$\boldsymbol{v}^t_{n,i} \in \mathcal{V}^t_n$} the $i$-th node that belongs to the navigational abstraction layer {$\mathcal{V}^t_n$} at time $t$.
Each observed motion event at a spatial location yields a sample on the orientation--speed cylinder {$\boldsymbol{z} = (\theta, \rho) \in [-\pi, \pi) \times \mathbb{R}^+$}, where $\theta$ is the heading direction and $\rho$ the motion speed. The collection of samples at each navigational node $v^{t}_{n,i}$ is modeled by a semi-wrapped Gaussian mixture:

\begin{equation}
    \label{eq:dynamics_gmm}
    p(\boldsymbol{z} \mid \boldsymbol{v}^{t}_{n,i}) = \sum_{k=1}^{K_i} \alpha^{}_k \; \mathcal{N}_{\text{sw}}\bigl(\boldsymbol{z} \,\big|\, \boldsymbol{\mu}^{}_k, \boldsymbol{\Sigma}^{}_k\bigr) \,,
\end{equation}

\noindent where {$K^{}_i$} is the number of mixture components at node $i$, {$\alpha^{}_k$} are mixing weights summing to one, and {$\mathcal{N}_{\text{sw}}$} denotes a semi-wrapped Gaussian of mean $\boldsymbol{\mu}$ and covariance $\boldsymbol{\Sigma}$ that accounts for the circular topology of $\theta$.
Because $\theta$ is periodic, a standard Gaussian centered near $\pm\pi$ would assign near-zero density to observations on the opposite side of the wraparound boundary.
The semi-wrapped construction remedies this by summing shifted replicas of the Gaussian over integer multiples of $2\pi$:

\begin{equation}\label{eq:semi_wrapped}
    \mathcal{N}_{\text{sw}}\bigl(\boldsymbol{z} \,\big|\, \boldsymbol{\mu}, \boldsymbol{\Sigma}\bigr) = \sum_{w=-W}^{W} \mathcal{N}(\boldsymbol{z} + (2\pi w, 0)^\top \mid \boldsymbol{\mu}, \boldsymbol{\Sigma}),
\end{equation}

\noindent with winding number {$W$} controlling the wrapping extent (typically $W{=}1$).

\subsection{Online Model Fitting}
\label{sec:model-architecture}

To support reasoning over dynamic environments, Rheos represents motion dynamics directly on top of a hierarchical 3DSG.

The formulation in~\cite{kucner2017enabling} processes all observed trajectories in batch: a full mean-shift clustering pass discovers modes from scratch, followed by Expectation-Maximization (EM) to fit the mixture.
This batch procedure has $\mathcal{O}(n^2_{} I)$ complexity (where $n$ is the sample count and $I$ the number of mean-shift iterations) and is designed for offline map construction from completed datasets.
Rheos adapts this pipeline for incremental, real-time operation through two key modifications: (i)~a fixed-capacity sample buffer with reservoir sampling, and (ii)~a BIC-based model order sweep with $K$-means++ initialization that replaces the expensive $\mathcal{O}(n^2)$ mean-shift clustering with a principled, linear-cost alternative.

Each spatial hash cell and, after binding, each dynamics node, maintains its own independent model (\cref{eq:dynamics_gmm}); all fitting operations described below are performed per-cell.

\textbf{Sample Buffer.} 
Unlike the work in~\cite{kucner2017enabling} which assumes that the complete set of trajectory observations is available at fitting time and stores all samples in memory, Rheos equips each cell with a fixed-capacity buffer of at most $M$ observations $\{\boldsymbol{z}\}_{j=1}^{M}$.
While the buffer is not full, incoming samples are simply appended.
Once $M$ is reached, each new sample replaces an existing entry with probability $M/T$ (where $T$ is the total number of observations seen so far), following Vitter's reservoir sampling algorithm~\cite{vitter1985random}.
This guarantees that the buffer remains a uniform random subsample of all observations ever received, bounding memory to $\mathcal{O}(M)$ per cell regardless of the observation horizon.

\textbf{BIC-Based Model Order Selection.}
A critical step in fitting a Gaussian mixture is choosing the number of components~$K^{}_i$ at each node $i$. For clarity, we will hereafter write $K$ instead of $K^{}_i$, assuming the node index is implicit.
The formulation in~\cite{kucner2017enabling}, relies on mean-shift clustering~\cite{cheng1995mean} with a Silverman-rule bandwidth to discover modes; the number of converged modes is then passed directly to EM as~$K$.
While intuitive, this approach couples model order to kernel bandwidth: a bandwidth that is too narrow over-segments the data (high~$K$), while one that is too wide collapses modes (low~$K$).

Rheos replaces this heuristic with a principled model selection sweep based on the Bayesian Information Criterion (BIC)~\cite{schwarz1978estimating}.
For each candidate component count $K \in \{1, \dots, K_{\max}\}$, a mixture with parameters $\hat{\Psi}_K = \{\alpha_k, \boldsymbol{\mu}_k, \boldsymbol{\Sigma}_k\}_{k=1}^{K}$ is initialized via $K$-means++ on the $(\theta, \rho)$~cylinder and refined by EM.
The BIC of the resulting model is:

\begin{equation}
    \label{eq:bic}
    \text{BIC}(K) = k^{}_p \ln n - 2\,\mathcal{L}(K) \,,
\end{equation}

where $n$ is the number of buffered observations, $\mathcal{L}(K) = \sum_{j=1}^{n} \ln p(\boldsymbol{z}_j \mid \hat{\Psi}_K)$ is the maximized semi-wrapped log-likelihood under the fitted parameters~$\hat{\Psi}_K$, and

\begin{equation}
    \label{eq:bic_params}
    k^{}_p = 6K - 1\,,
\end{equation}
\noindent is the number of free parameters (two mean entries, three independent covariance entries, and one mixing weight per component, minus one global constraint $\sum_k \alpha_k = 1$).
The number of mixture components at each node~$i$ is then selected independently as
$K_i = \argmin_{K} \text{BIC}(K)$,
balancing goodness of fit against model complexity without dependence on a bandwidth parameter.

The $K$-means++ initialization~\cite{arthur2006k} samples the first center uniformly from the data, then selects each subsequent center with probability proportional to the squared circular-linear distance to its nearest existing center, yielding well-separated initial seeds.
Labels are assigned by nearest center under the same metric, and EM proceeds as described below.

\textbf{Expectation-Maximization.} 
Starting from the initial cluster assignments, EM iteratively refines the component parameters $\{\alpha^{}_k, \boldsymbol{\mu}^{}_k, \boldsymbol{\Sigma}^{}_k\}_{k=1}^K$ by maximizing the semi-wrapped log-likelihood~(\cref{eq:dynamics_gmm}).
In the \emph{E-step}, each sample $\boldsymbol{z}^{}_j$ is soft-assigned to each component~$k$ and winding copy~$w$ in proportion to the posterior responsibility:

\begin{equation}
    \label{eq:posterior-responsibility}
    r^{}_{jkw} \propto \alpha^{}_k \,\mathcal{N}(\boldsymbol{z}^{}_j + (2\pi w, 0)^\top \mid \boldsymbol{\mu}_k, \boldsymbol{\Sigma}^{}_k) \,,
\end{equation}

In the \emph{M-step}, the responsibilities are used to re-estimate each component's weight ($\alpha^{}_k$), mean ($\boldsymbol{\mu}^{}_k$), and covariance ($\boldsymbol{\Sigma}^{}_k$).
Covariance matrices are regularized via diagonal clamping and a Cauchy-Schwarz off-diagonal bound to ensure positive-definiteness on the cylinder.

To reduce computational cost, Rheos pre-computes each covariance inverse $\boldsymbol{\Sigma}_k^{-1}$ and its normalizing constant $|\boldsymbol{\Sigma}_k|^{-1/2}$ once per \emph{M-step}, and reuses them across all $n \times (2W{+}1)$ \emph{E-step} evaluations. EM terminates when the log-likelihood change falls below a threshold or a maximum number of iterations is reached.

\textbf{Complexity.}
Each model update runs the full BIC sweep: for each candidate $K \in \{1, \dots, K_{\max}\}$, $K$-means++ initialization costs $\mathcal{O}(MK)$ and EM costs $\mathcal{O}(MK(2W{+}1)I^{}_{\text{em}})$ per iteration, yielding a total per-cell cost of $\mathcal{O}(K_{\max} \cdot M \cdot K_{\max} \cdot (2W{+}1) \cdot I^{}_{\text{em}})$, where $M$ is the buffer capacity and $W$ the winding number.
Since $K_{\max}$, $W$, and $I^{}_{\text{em}}$ are small bounded constants in practice ($K_{\max}{=}5$, $W{=}1$, $I^{}_{\text{em}} \leq 100$), this reduces to $\mathcal{O}(M)$ per cell, making each update linear in the buffer size.
Compared with the batch $\mathcal{O}(M^2 I)$ mean-shift initialization of~\cite{kucner2017enabling}, the $K$-means++ seeding used by the BIC sweep reduces the initialization cost from quadratic to linear in the number of samples.

\subsection{System Integration}

Rheos adopts the real-time integration architecture of Aion~\cite{catalano2025aion}: scene graph updates and agent detections are processed asynchronously, per-node model updates are parallelized across nodes, and motion dynamic computations are exposed via ROS services compatible with Hydra~\cite{hughes2022hydra}.
\section{Experimental Evaluation}
\label{sec:experiments}

\begin{figure}[t]
    \centering
    \subfloat[Top-down view\label{fig:simulation-top-view}]{
        \includegraphics[width=0.27\linewidth, angle=90]{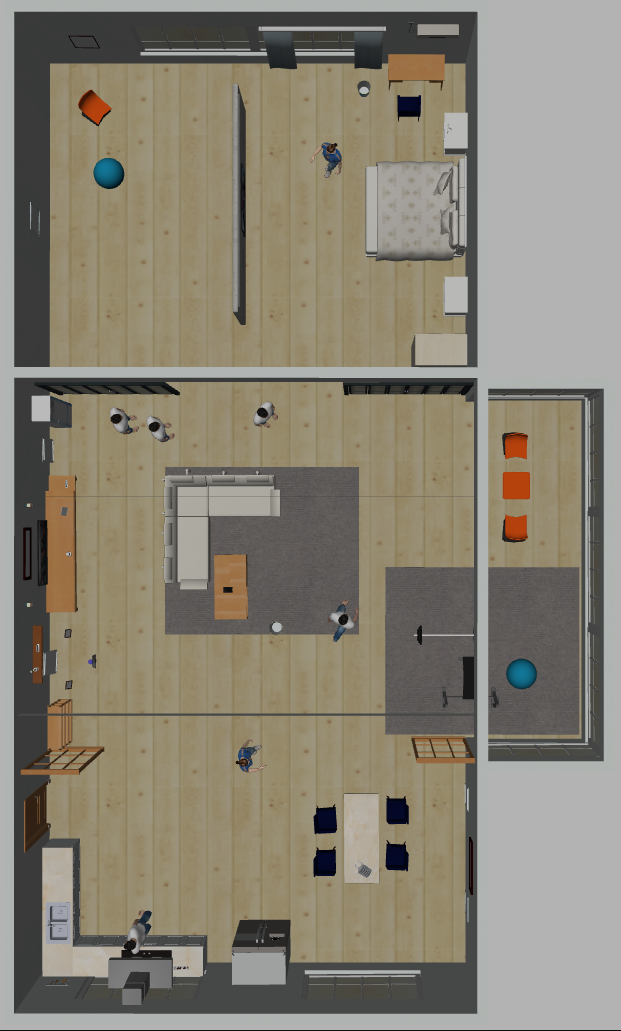}
    }\hfill
    \subfloat[Perspective view\label{fig:simulation-perspective-view}]{
        \includegraphics[width=0.48\linewidth]{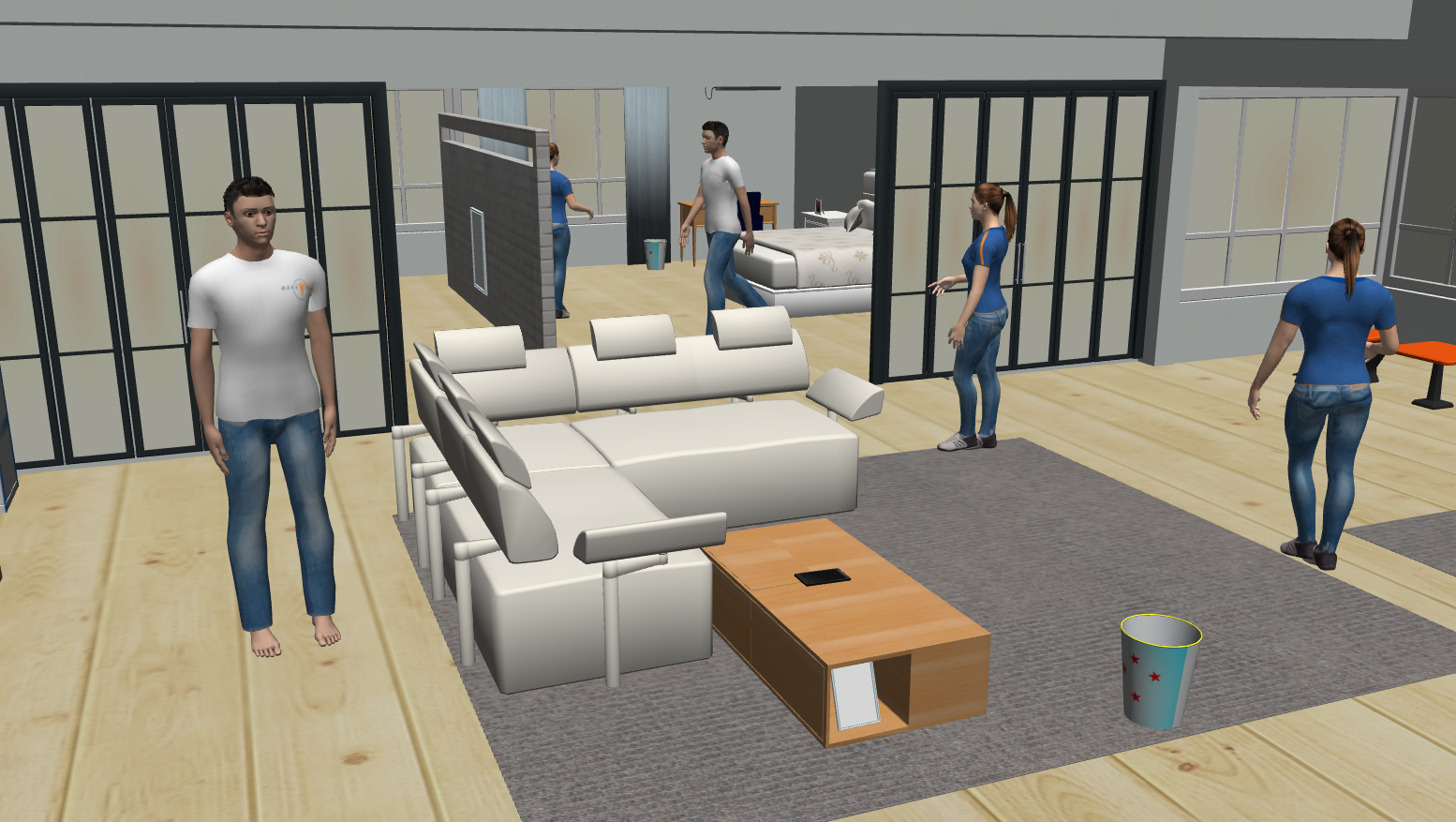}
    }
    \caption{\textbf{Overview of the simulation environment} shown from a top-down view (left) and a perspective view (right).}
    \label{fig:simulation}
\end{figure}

Our experiments are designed to evaluate three aspects of Rheos: (1)~how accurately the continuous SW-GMM captures directional motion patterns compared to the discrete histogram baseline, (2)~how the spatial resolution of the navigational layer affects prediction quality and computational cost, and (3)~how computational cost and memory scale with resolution during online operation.

\subsection{Simulation Environment}
\label{sec:simulation}

The environment used for evaluation is a house scenario built in Gazebo using AWS RoboMaker Small House World, combined with a customized version of PedSim simulator~\cite{xie2021towards} extended to include this setting (see \cref{fig:simulation}). Additionally, the agent behaviors were modified to define movement patterns and queuing behaviors at key locations. The environment measures approximately 18\,m by 10\,m and contains 7 agents moving following realistic human dynamics, simulated in PedSim through a social-force model~\cite{helbing1995social}. We use the simulator-provided detections to compute the dynamic model discussed in~\cref{sec:methodology}.

\subsection{Metrics}

Since model parameters are fixed at evaluation time, we assess predictive performance using scoring rules rather than likelihood functions in the estimation sense. The reported metrics therefore quantify how much probability mass the learned distribution assigns to held-out observations, enabling comparison between continuous and histogram-based representations.

To evaluate predictive performance without introducing discretisation artifacts, we report the \emph{Mean Log Predictive Density (MLPD)}, which measures the log probability density assigned to the exact observed direction. For comparability with prior histogram-based motion modelling work, we additionally report the \emph{Mean Predictive Probability (MPP)} metric used in \cite{verdoja2024bayesian}. This metric discretizes the angular domain into $B$ bins and evaluates the probability mass assigned to the observed bin.
We include both metrics because they serve complementary roles: MPP enables direct comparison with prior histogram-based work, while MLPD, as a strictly proper scoring rule, reveals the inherent advantage of continuous representations free from discretisation artifacts.

\textbf{Mean Log Predictive Density.}
To eliminate discretisation bias, we evaluate the marginal directional density at each observation's exact angle:
\begin{equation}
    \text{MLPD}(O) = \frac{1}{N} \sum_{j=1}^{N} \log p(\theta^{}_j \mid d^{}_{c^{}_j}) \,.
    \label{eq:cont_loglik}
\end{equation}
This corresponds to the average log predictive density under a fixed model, penalizing overconfident incorrect predictions and enabling resolution-independent comparison between continuous and discrete representations.

For Rheos, the marginal $p(\theta \mid d^{}_c)$ is computed analytically from the fitted GMM by marginalising the speed dimension.
For the histogram-based baseline, the density is piecewise-constant: $p(\theta\mid d^{}_c) = P(\theta\mid d^{}_c)^{}_b / \Delta\theta$, where $P(\theta\mid d^{}_c)^{}_b$ is the normalised bin probability containing $\theta$, and $\Delta\theta = 2\pi/B$.

\textbf{Mean Predictive Probability.}
Given a set $O = \{(\boldsymbol{p}^{}_j, \theta^{}_j)\}_{j=1}^N$ of observed positions and corresponding discretized motion directions, the MPP of a model $\mathcal{D}$ is defined as:
\begin{equation}
    \text{MPP}(O \mid \mathcal{D}) = \frac{1}{N} \sum_{j=1}^{N} P(\theta^{}_j \mid d^{}_{c^{}_j}) \,,
    \label{eq:avg_likelihood}
\end{equation}
where $c^{}_j$ is the grid cell containing observation $\boldsymbol{p}^{}_j$, and $P(\theta^{}_j \mid d^{}_{c_j})$ is the probability assigned by the model to the observed direction at that cell.
This metric measures the average probability mass assigned to the observed directions. While intuitive and directly comparable to histogram-based baselines, it remains sensitive to angular discretisation.

For Rheos, since it models directions as a continuous density, we compute the bin probability $P(\theta^{}_j \mid d^{}_{c^{}_j})$ by integrating the GMM over the corresponding angular bin interval, ensuring that the full probability mass within each bin is captured.
This procedure removes the ability of the model to represent fine angular structure and effectively reduces the prediction to a histogram approximation. As a result, the metric aligns more naturally with Aion’s discrete representation.

\subsection{Comparison Framework}
\label{sec:comparison-framework}

\textbf{Reference MoD Construction.}
To establish an empirical upper bound on prediction quality, a reference MoD is constructed directly from the complete set of observed trajectories, following the procedure in~\cite{verdoja2024bayesian}.
The environment is discretized into a grid at resolution $\delta=0.1$\,m, finer than any model under evaluation, and each cell accumulates a histogram of motion directions over $B$ angular bins.
The resulting categorical distributions are normalized to produce a probability grid $\mathcal{D}=\{d^{}_c | c\in(1,\dots,C)\}$, where $d^{}_c$ is the $B$-bin distribution at cell $c$.
Because this reference map has access to the full training data and uses a finer spatial discretization than either Rheos or Aion, it represents the best achievable performance for the given observations; the \emph{reference MPP} is evaluated on its own training data, providing the upper bound of the metric.

\textbf{Metrics Variants}
We report two variants of the metrics described above.
The \emph{overall} probability evaluates all reference trajectory points. Since this makes a highly accurate but spatially sparse model pulled toward the uniform baseline, we additionally report the \emph{covered-only} probability which restricts evaluation to the subset of trajectory points that fall in cells where the model has data, isolating prediction quality from coverage extent.

\textbf{Generalization.}
To assess how well the learned models predict \emph{unseen} motion patterns, we train both systems on one simulation run and evaluate on a second run of the same scenario.
Because the simulation is stochastic, the two runs share the same macroscopic motion patterns but differ in individual trajectories, providing a straightforward train/test split.

\textbf{Spatial Resolution.}
Both systems inherit their spatial resolution from the underlying 3DSG navigational layer.
Hydra~\cite{hughes2022hydra} generates navigational nodes by recursively decomposing 2D surface clusters along their principal axis until each resulting region falls below a configurable size threshold.
Consequently, varying this threshold controls the density of navigational nodes and, by extension, the spatial granularity of the dynamics layer: a smaller threshold produces more, finer-grained nodes.
The evaluation grid resolutions~$\delta$ in~\cref{tab:likelihood} refer to the cell size of the comparison grid onto which both predicted distributions are projected, matching the density of the navigational nodes.

\subsection{Results and Analysis}

We evaluate Rheos against Aion~\cite{catalano2025aion}, a motion modelling system integrated into the same 3DSG framework. Both systems operate on the same navigational layer of the scene graph and receive identical detection inputs, isolating the effect of the dynamics model. We adopted the default configuration for Aion and repeated the evaluation across four spatial resolutions ($\delta \in \{0.2, 0.3, 0.5, 1.0\}$\,m) to characterize the resolution--performance trade-off.

Additionally we consider a \emph{uniform model} with probability $P(\theta) = 1/B$ at every cell, representing the prediction quality of a model with no directional information.
The reference MPP serves as the empirical upper bound, and the uniform model as a na\"ive baseline.

\textbf{Mean Log Predictive Density.}
Under the MLPD metric (see \Cref{tab:continuous}), both systems score below the uniform baseline ($-1.84$) at all resolutions.
This result arises from the logarithmic scoring rule: both models concentrate probability mass in the observed preferred motion directions at each cell; when test observations fall in low-density regions the resulting large negative log-densities dominate the arithmetic mean.
The uniform density avoids such extremes by spreading mass evenly across all angles.

However, at all resolutions Rheos consistently outperforms Aion. The gap is largest at $\delta{=}0.2$\,m ($-2.65$ vs.\ $-6.70$), where histogram sparsity is most severe and widens as Aion's angular resolution increases from $B{=}8$ to $B{=}360$: the histogram becomes sparser, many bins contain zero probability, and the piecewise-constant density collapses to the numerical floor at unobserved angles, worsening its score.
Rheos's continuous GMM is resolution-independent by construction, providing smoother tails and thus more robust density estimates at arbitrary query angles.


\begin{table}
    \centering
    \caption{\textbf{Mean Log Predictive Density at four spatial resolutions} (direction only, higher is better). Reference (upper bound): $-0.54$; Uniform baseline: $-1.84$.}
    \label{tab:continuous}
    \begin{tabularx}{\columnwidth}{c >{\centering\arraybackslash}X >{\centering\arraybackslash}X}
    \toprule
    $\delta$ [m] & Rheos & Aion \\
    \midrule
    0.2 & $\mathbf{-2.65}$ & $-6.70$ \\
    0.3 & $\mathbf{-2.17}$ & $-2.37$ \\
    0.5 & $\mathbf{-1.91}$ & $-2.23$ \\
    1.0 & $\mathbf{-1.85}$ & $-2.16$ \\
    \bottomrule
    \end{tabularx}
\end{table}

\textbf{Mean Predictive Probability.} 
\Cref{tab:likelihood} reports the MPP at each resolution with $B{=}8$ angular bins. It is important to note that the MPP metric inherently favors discrete histogram representations such as Aion’s. Because the metric evaluates probability mass within a fixed set of angular bins, continuous models must first integrate their density over the corresponding bin intervals. This discretisation removes the main advantage of Rheos (its ability to represent smooth directional densities) and effectively reduces the continuous model to a histogram approximation during evaluation. However, despite the information loss introduced, Rheos outperforms state-of-the-art methods at every resolution.
As resolution coarsens, the overall probabilities converge toward the uniform baseline ($0.125$) since most observations fall in cells without model data.

The \emph{overall} and \emph{covered-only} scores illustrate an important distinction: the overall metric penalizes spatial sparsity by assigning the uniform density to uncovered cells, while the covered-only metric isolates the quality of the model \emph{where it has data}.
Since both systems cover only a fraction of the observation grid, the overall metric is heavily influenced by the uniform fallback at all resolutions.


\begin{table}
    \centering
    \caption{\textbf{Mean Predictive Probability (MPP)} ($B{=}8$ bins) \textbf{at four spatial resolutions, $\delta$}. Rheos probabilities are computed by integrating the GMM over each angular bin. ``Overall'' assigns $1/B{=}0.125$ to uncovered cells; ``Covered'' restricts evaluation to cells with model data. Reference (upper bound): $0.600$; Uniform baseline: $0.125$.} 
    \label{tab:likelihood}
    \begin{tabularx}{\columnwidth}{c l *{4}{>{\centering\arraybackslash}X}}
        \toprule
        $\delta$ &  & \multicolumn{2}{c}{Overall} & \multicolumn{2}{c}{Covered-only} \\
        \cmidrule(lr){3-4}\cmidrule(lr){5-6}
        {[m]} &  & Rheos & Aion & Rheos & Aion \\
        \midrule
        0.2 & & \textbf{0.322} & 0.276 & \textbf{0.492} & 0.376 \\
        0.3 & & \textbf{0.218} & 0.155 & \textbf{0.537} & 0.455 \\
        0.5 & & \textbf{0.164} & 0.150 & \textbf{0.589} & 0.416 \\
        1.0 & & \textbf{0.135} & 0.131 & \textbf{0.340} & 0.255 \\
        \bottomrule
    \end{tabularx}
\end{table}

\textbf{Runtime and Memory Across Resolutions.}
\Cref{tab:runtime_resolution} reports per-iteration runtimes and memory footprints across all tested spatial resolutions.
Detection association cost scales linearly with the number of navigational nodes: at $\delta{=}0.2$\,m with ${\approx}535$ navigational nodes, Rheos averages $42.4$\,ms per callback; at $\delta{=}1.0$\,m with ${\approx}77$ nodes, this drops to $1.6$\,ms.
Model fitting cost ranges from $5.1$\,s to $7.1$\,s across resolutions because the BIC sweep evaluates $K_{\max}{=}5$ candidate models per cell, each requiring a full EM run.
The cost increases slightly at coarser resolutions because fewer, larger cells accumulate more diverse observations, leading to mixtures with more components that require more EM iterations.
Rheos's fitting cost is substantially higher than Aion's because Aion only accumulates histogram bin counts ($\mathcal{O}(1)$ per observation), whereas Rheos runs iterative EM optimization per cell; this added cost is the price of obtaining continuous, resolution-independent densities with explicit uncertainty, as reflected in the MLPD and MPP improvements.
However, model updates are performed asynchronously at configurable intervals ($10$\,s in our setup),
ensuring that the dynamics layer remains current during online operation.

Memory footprint scales proportionally with the number of nodes, from $328.8$\,KB at $\delta{=}0.2$\,m to $171.8$\,KB at $\delta{=}1.0$\,m.

\begin{table*}
    \centering
    \caption{\textbf{Per-iteration runtime} (mean $\pm$ std) and \textbf{in-memory footprint} (in kB) across four spatial resolutions. ``Det. Assoc.'' = detection association callback (ms); ``Model'' = dynamics model fit (ms). All measurements from online operation on ${\approx}32$\,k observations.}
    \label{tab:runtime_resolution}
    \setlength{\tabcolsep}{4pt}
    \begin{tabular*}{\textwidth}{@{\extracolsep{\fill}}ccccccccc}
    \toprule
     & & \multicolumn{2}{c}{Det.\ Assoc.\ [ms]} & \multicolumn{2}{c}{Model Update [ms]} & \multicolumn{2}{c}{Memory [kB]} \\
    \cmidrule(lr){3-4}\cmidrule(lr){5-6}\cmidrule(lr){7-8}
    $\delta$ [m] & Nodes & Rheos & Aion & Rheos & Aion & Rheos & Aion \\
    \midrule
    0.2 & ${\approx}535$ & $42.4 \pm 26.0$ & $100.5 \pm 48.4$ & $5061 \pm 1764$ & $140 \pm 55$ & 328.8 & 215.8 \\
    0.3 & ${\approx}260$ & $14.6 \pm \phantom{0}8.9$ & $\phantom{0}12.4 \pm \phantom{0}6.6$ & $5664 \pm 1842$ & $\phantom{0}37 \pm 11$ & 250.3 & 113.8 \\
    0.5 & ${\approx}160$ & $\phantom{0}5.1 \pm \phantom{0}3.2$ & $\phantom{0}13.0 \pm \phantom{0}6.8$ & $6833 \pm 2176$ & $\phantom{0}30 \pm 10$ & 201.8 & 106.3 \\
    1.0 & $\phantom{0}{\approx}77$ & $\phantom{0}1.6 \pm \phantom{0}0.9$ & $\phantom{00}4.3 \pm \phantom{0}2.2$ & $7066 \pm 2178$ & $\phantom{0}20 \pm \phantom{0}9$ & 171.8 & \phantom{0}91.6 \\
    \bottomrule
    \end{tabular*}
\end{table*}

\begin{figure}[t]
    \centering
    
    \subfloat[Rheos\label{fig:qualitative-rheos}]{
        \includegraphics[width=\linewidth]{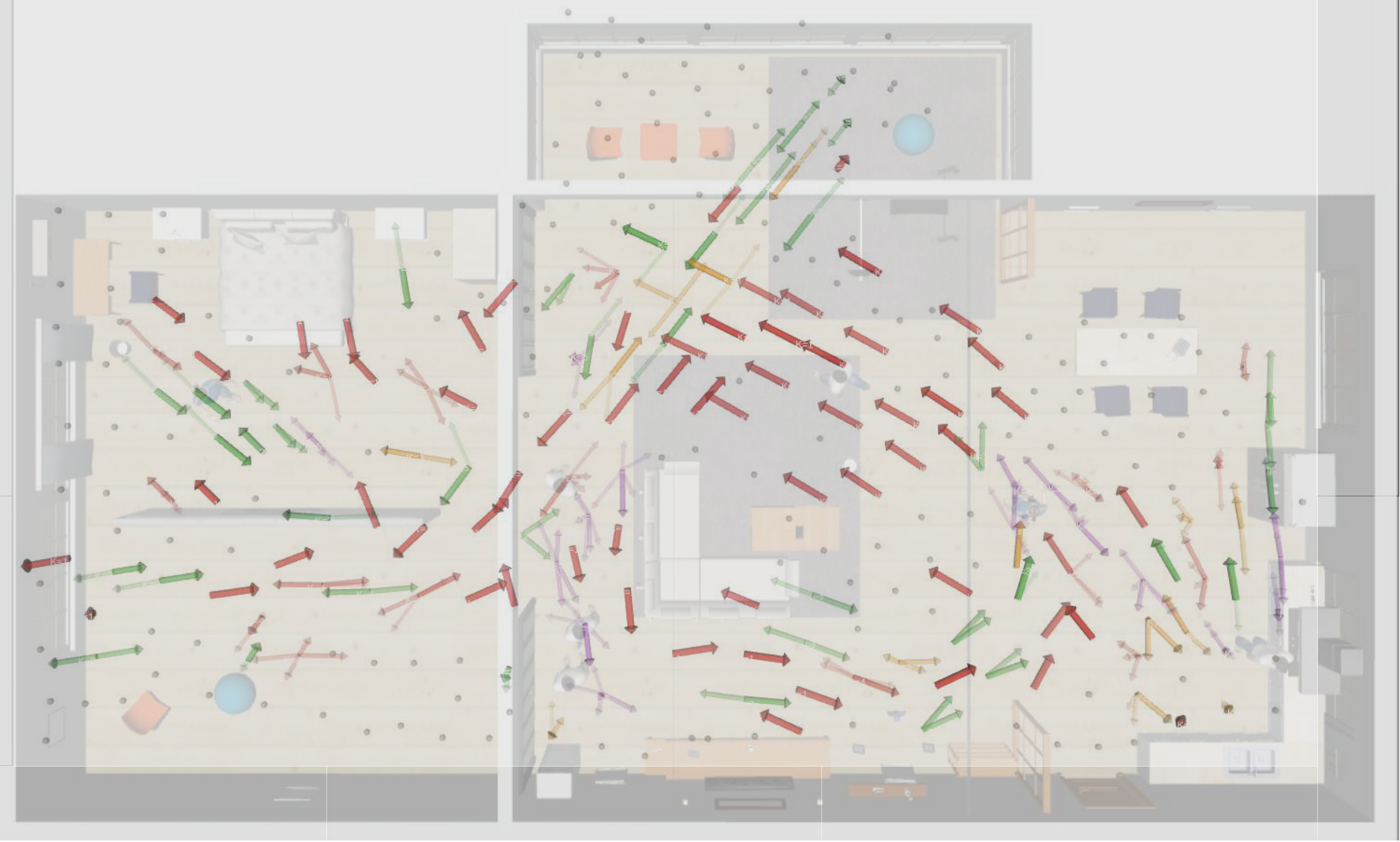}
    }
    \hfill
    \subfloat[Aion\label{fig:qualitative-aion}]{
        \includegraphics[width=\linewidth]{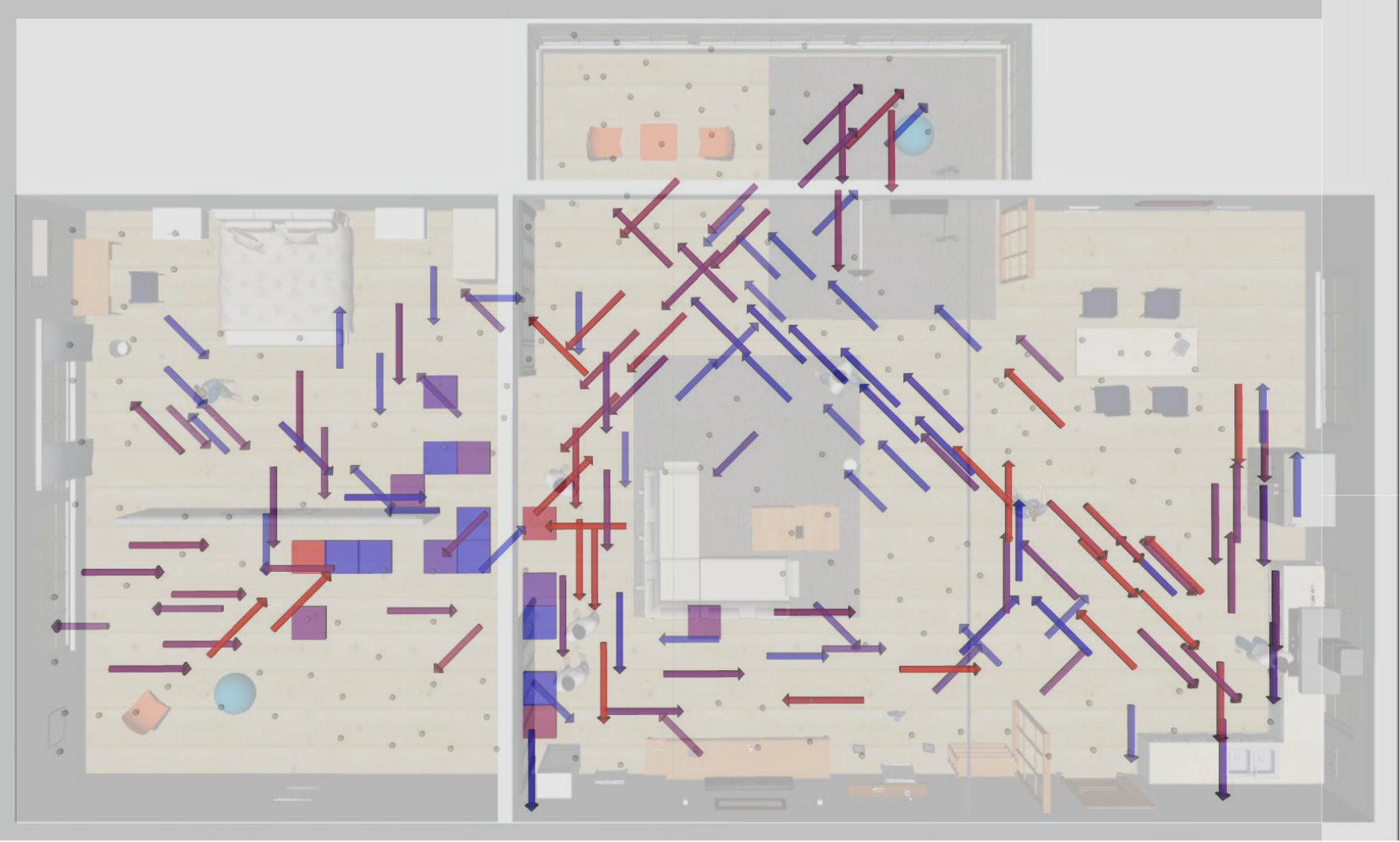}
    }
    \caption{\textbf{Qualitative comparison of motion dynamics} reconstructed for the same scene from Rheos (top) and Aion (bottom). Arrow markers indicate the dominant direction of motion. While both methods capture the broad reach of movement, Rheos reveals finer motion structure, producing a more granular and continuous map of motion patterns.}
    \label{fig:qualtative-comparison}
\end{figure}

\textbf{Qualitative Visualization.} The qualitative comparison in~\cref{fig:qualtative-comparison} reveals that while Aion is characterized by high directional variance, capturing expansive and less constrained movement patterns throughout the environment, Rheos generates a more structured and aligned vector field, effectively delineating fluid pedestrian motion and established navigational corridors.

\subsection{Ablation: Model Order Selection}
\label{sec:ablation}

To isolate the contribution of the BIC-based model order selection introduced in~\cref{sec:model-architecture}, we compare the proposed BIC sweep against the original mean-shift initialization from~\cite{kucner2017enabling} within the same Rheos framework.
Both configurations share the same reservoir sampling buffer, EM refinement, and system integration; only the method for choosing~$K_i$ differs.
We run both methods on the same simulation at $\delta{=}0.5$\,m with identical parameters ($M{=}200$, $W{=}1$, $K_{\max}{=}5$ for BIC).

\begin{table}[t]
    \centering
    \caption{\textbf{Ablation: BIC sweep vs.\ Mean-Shift initialization} at $\delta{=}0.5$\,m.
    Both methods run within the same Rheos framework; only the model order selection differs.}
    \label{tab:ablation}
    \begin{tabularx}{\columnwidth}{Xcc}
        \toprule
         & Mean-Shift & BIC Sweep \\
        \midrule
        Active nodes          & 158         & 140 \\
        Mean $K$              & 1.11        & \textbf{2.17} \\
        \midrule
        MLPD (overall)        & $-2.00$     & $\mathbf{-1.93}$ \\
        MLPD (covered)        & $-3.76$     & $\mathbf{-2.87}$ \\
        MPP  (overall)        & $0.139$     & $\mathbf{0.164}$ \\
        MPP  (covered)        & $0.297$     & $\mathbf{0.589}$ \\
        \midrule
        Update time [s]       & $4.8 \pm 3.8$ & $6.8 \pm 2.2$ \\
        \bottomrule
    \end{tabularx}
\end{table}

Mean-shift initialization produces $K{=}1$ in 92\% of cells (146 of 158), confirming that the Silverman-rule bandwidth collapses multimodal flows into single components.
BIC sweep selects a richer model order (mean $K{=}2.17$), with 40\% of cells at $K{=}1$ and the remainder distributed across $K{=}2$--$5$.
This directly improves predictive quality: the covered-only MLPD improves from $-3.76$ to $-2.87$, and the covered-only MPP nearly doubles from $0.297$ to $0.589$ (\cref{tab:ablation}).
The cost is a ${\approx}1.4{\times}$ increase in model update time ($6.8$\,s vs.\ $4.8$\,s), as the BIC sweep evaluates $K_{\max}{=}5$ candidates per cell; this remains within the configurable $10$\,s update interval.
\section{Conclusion}\label{sec:conclusion}

In this paper we presented Rheos, a system that augments hierarchical 3D scene graphs with continuous directional motion models, unifying Maps of Dynamics (MoDs) with semantically structured spatial representations.
By replacing discrete histograms with semi-wrapped Gaussian mixtures at each navigational node, Rheos provides resolution-independent directional distributions with explicit uncertainty while maintaining periodic model updates within configurable intervals, suitable for online robotic operation.

Our evaluation across four spatial resolutions shows that Rheos consistently outperforms the discrete histogram baseline under the continuous Mean Log Predictive Density (MLPD) metric, with the largest gains at fine resolutions where histogram sparsity is most pronounced. Under the discrete Mean Predictive Probability (MPP) metric, which structurally favors histogram representations, Rheos still outperforms the histogram baseline at every resolution, demonstrating that principled model order selection produces mixtures that assign higher probability mass to the correct angular bins even under unfavorable evaluation conditions. These results confirm that continuous mixture models are a more principled foundation for directional flow representation in scene graphs, without sacrificing compatibility with existing discrete evaluation frameworks.

We provide a complete open-source implementation integrated into the Hydra 3DSG framework, alongside a simulation-based benchmark and evaluation suite. Future work will focus on incorporating temporal prediction of motion patterns into the continuous representation, coupling flow estimates with downstream planners for human-aware navigation, and validating the approach in larger real-world environments where non-stationary dynamics and longer observation horizons can stress-test the incremental fitting pipeline.


\section{Acknowledgment}\label{sec:acknowledgment}
The authors acknowledge the use of Claude Opus 4.6~\cite{anthropic2026claude} for improving readability across all sections, as well as generating documentation and improving code readability in the project.

\balance
\bibliographystyle{unsrt}
\bibliography{bibliography}

\end{document}